\documentclass[twoside,leqno,twocolumn]{article}  
\usepackage{ltexpprt} 
\usepackage[cmex10]{amsmath}
\usepackage{array}
\usepackage{url}
\usepackage{amssymb} 
\usepackage{bm}
\usepackage{setspace}
\usepackage{graphicx}
\usepackage{enumerate}
\usepackage{algorithm}
\usepackage{algpseudocode}
\usepackage{comment}

\newtheorem{thm}{Theorem}
\newtheorem{lem}{Lemma}

\newtheorem{cor}{Corrolary}
\newtheorem{definition}{Definition}[section]
\newcommand{\eat}[1]{}

\newcommand{\tab}{\hspace*{2em}}
\newcommand{\mc}[1]{\mathcal{#1}}
\newcommand{\tup}[1]{\textup{#1}}
\begin{document}

\title{Uncovering Group Level Insights with Accordant Clustering}

\author{Amit Dhurandhar\thanks{IBM Research, adhuran@us.ibm.com} \and
Margareta Ackerman\thanks{San Jose State University, margareta.ackerman@sjsu.edu} \and
Xiang Wang\thanks{Google, xiangwa@google.com. Xiang contributed to this work while at IBM Research}
}
\date{}

\maketitle


\begin{abstract} \small \baselineskip=9pt 
Clustering is a widely-used data mining tool, which aims to discover partitions of similar items in data. We introduce a new clustering paradigm, \emph{accordant clustering}, which enables the discovery of (predefined) group level insights. Unlike previous clustering paradigms that aim to understand relationships amongst the individual members, the goal of accordant clustering is to uncover insights at the group level through the analysis of their members. Group level insight can often support a call to action that cannot be informed through previous clustering techniques. We propose the first accordant clustering algorithm, and prove that it finds near-optimal solutions when data possesses inherent cluster structure. The insights revealed by accordant clusterings enabled experts in the field of medicine to isolate successful treatments for a neurodegenerative disease, and those in finance to discover patterns of unnecessary spending. 

\end{abstract}


\section{Introduction}

As one of the most fundamental data mining tools, clustering is employed in a variety of domains that span from biology~\cite{clustbio} to marketing~\cite{clustmarket}, applicable in nearly all disciplines where data is utilized. The ubiquity of clustering is largely a result of its general and deceptively simple aim to discover partitions of similar items in data. Unfortunately, this aim is inherently ambiguous, as the same dataset can often be clustered in multiple meaningful ways. As such, the utility of any given clustering is application-dependent. The goal of clustering then is, not only to uncover meaningful cluster structure but, to find a partitioning that is useful for the application at hand.

The need for selecting among meaningful clusterings arises when we wish to uncover group level insight. For example, medical professionals may aim to gain insight into the performance of competing treatments through the analysis of several treatment groups. In this case, the goal is to discover not only which treatments are more effective, but also the demographics for which they are best suited. As such, we may wish to cluster patients across all treatments groups based on both demographic data (such as age, race, and gender) and the results of the treatment. While there may be several ways to meaningfully partition the patient data, \emph{not all high quality clusterings will help differentiate among the treatments}. For example, a clustering in which patients within treatment groups are evenly distributed across the resulting clusters may not be helpful for this application as no actionable conclusion can be drawn about the efficacy of the treatments in relation to the demographics. Instead, we are looking for a (high quality) clustering in which (at least some of the) clusters contain a significant proportion of one or more treatment groups. 

 Such a clustering will enable medical professionals to uncover relationships within and between the treatment groups by identifying similar characteristics amongst a significant portion of their members.\eat{\footnote{See Section~\ref{exp} for experiments on medical data and additional examples.}} We may find that a certain cluster (say, young women), containing a significant proportion of the first two treatment groups, responded well to the first, but not to the second, treatment. In a similar way, another cluster may reveal that a different demographic (say, older men) exhibit a harmful side-effect on the third treatment. Such insight may result in a call to action placing additional resources into promising treatments and/or terminating risky ones.

Despite the vast number of proposed frameworks, existing clustering paradigms focus on discriminating between individual instances, without taking into account the relationships amongst their underlying groups. Supervised and semi-supervised frameworks allow user input to help identify a meaningful partitioning of the individual members, but are no better than classical methods for discovering group insights (see the next section for a more details). 


In order to discover meaningful relationships within and between groups, we propose the notion of \emph{accordant clustering,} where sufficiently many elements in the same group are in ``accordance'' with respect to their cluster assignment. In this setting, the purpose of individual instances is to represent their underlying groups. The objective of accordant clustering is to balance two distinct aims, (1) discovering  inherent structure in data, an objective it shares with all other clustering paradigms, and (2) to combine elements that belong to the same group while minimizing violations to the first objective. The combination of these objectives allows accordant clustering to discover meaningful clusterings that support the discovery of insights at the group level.
\textbf{\begin{figure*}[t]
\vspace*{-3em}
\centering
\begin{tabular}{ cc }
	\includegraphics[width=130px, height=130px]{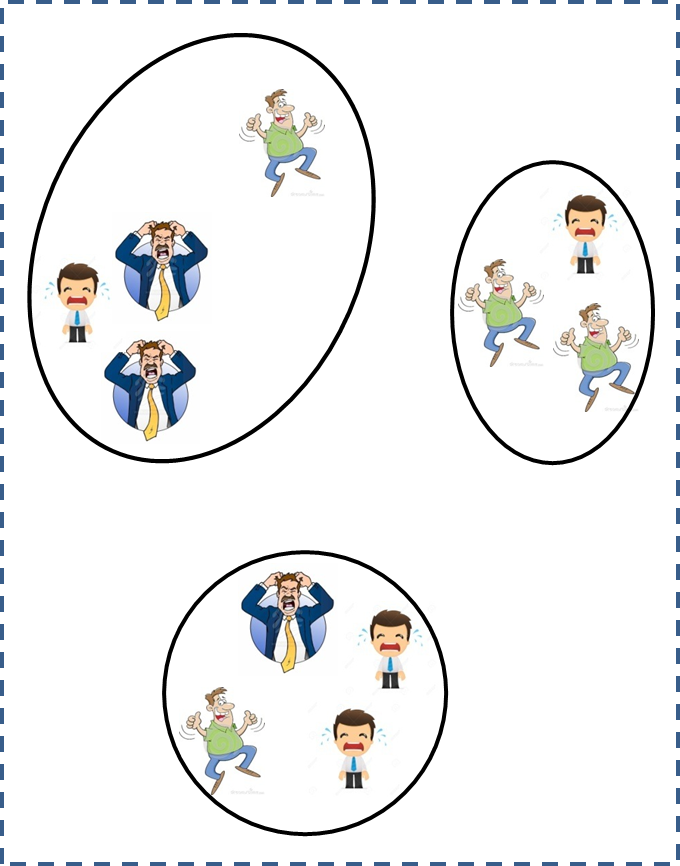}
	& \includegraphics[width=130px, height=130px]{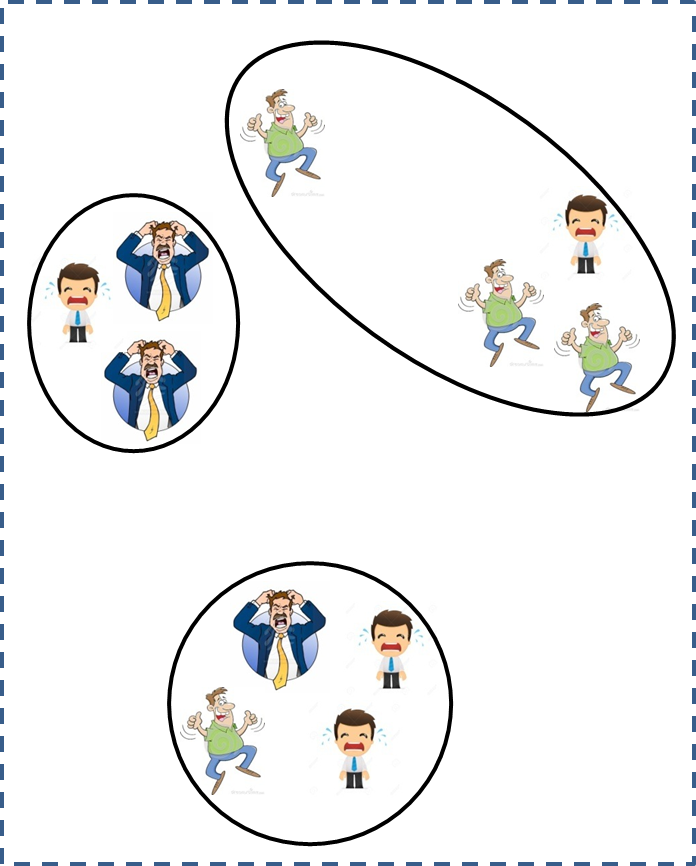}\\
	{Not accordant} 
	& {Accordant Clustering}
\end{tabular}
\caption{Two clusterings of a dataset consisting of three groups (happy, sad, and angry) are shown. Both clusterings represent inherent structure in data. However, given a threshold ratio of $0.75$, only the clustering on the right is accordant, as it contains a cluster containing three quarters of the happy group. As such, the clustering on the right is ($1$,$0.75$)-accordant.}
\label{fig:ac:intro}
\end{figure*}}
In addition to introducing this new paradigm, we propose the first accordant clustering algorithm, which is based on the popular $k$-means method. We begin with a formal analysis of our algorithm, by first showing that it converges as well as uncovers provably near-optimal solutions when data possesses inherent cluster structure. We then report results from two real domains, where the clusterings produced by our method enabled experts to ascertain actionable insight. Lastly, we  report results on six UCI datasets showing that our method finds higher quality accordant clusterings relative to its adapted competitors.

\section{New Clustering Framework} 



We now introduce a formal framework for accordant clustering which enables group level insights.

\subsection{Formal Framework and Definitions}
\label{fw}

Let the input dataset $X \subset R^n$ be the union over $m$ groups, such that $X = \{X_1 \cup \cdots \cup X_m\}$. A clustering of $X$, assuming $k$ clusters, is denoted $\mc{C} = \{C_1, \dots, C_k\}$. The proportion of elements in a group $X_i$ that are clustered together are given by the vector $ f = \frac{1}{|X_i|} \left[~|C_1 \cap X_i|,~...,~|C_k \cap X_i|~\right]$. Clustering $\mc{C}$ is \emph{$t$-accordant on $X_i$} if $\exists$ $j\in\{1,...,k\}$ such that the $j^{th}$ component of $f$, $f_j \geq t$. We now introduce our main definition. 

\begin{definition}$(r,t)$-accordant clustering:\label{tr:ac}
\hspace{-1mm} Given a set $X$ subdivided into $m$ groups $\{X_1 \cup \cdots \cup X_m\}$, a clustering $\mc{C}$ of $X$ is \emph{$(r,t)$-accordant} if there exist at least $r$ distinct groups on which $\mc{C}$ is $t$-accordant.
\end{definition}

$C$ is the \textbf{optimal $(r,t)$-accordant clustering} with respect to objective function $\phi$ if it attains the best cost among all  $(r,t)$-accordant clusterings. That is, $C = argmin_{C} \{\phi(C) \mid C \textrm{ is an }(r,t)\textrm{-accordant clustering}\}.$ 



Figure~\ref{fig:ac:intro} depicts an example of a $0.75$-accordant clustering (defined explicitly as $(1,0.75)$-accordant). The constraints of Definition~\ref{tr:ac} emphasize the fact that gaining insight into relationships amongst the groups depends strongly on clusters representing a substantial proportion of their data. If $t=0.75$, i.e., we want some cluster to have $75$\% or more instances from one of the 3 groups (happy, sad and angry) depicted, then the right hand side clustering would be accordant since it has 3 out of the 4 happy people belonging to a cluster. The left hand side clustering is what would be obtained for its unsupervised counterpart. We see here that the accordant clustering is obtained for a slight penalty based on an objective $\Phi(.)$, that the clustering algorithms try to minimize. Hence, if $\mc{C}$ is the accordant clustering, $\Phi(\mc{C})$ would indicate its quality.

\textbf{Feasibility:} Of course, given $r$ and $t$, $k$ cannot be arbitrarily large to obtain an accordant clustering. We thus have the following result regarding feasibility.

\begin{lem}
Given a dataset $X$ of size $N$ partitioned into $m$ groups, with $n_1,...,n_r$ being the sizes of the $r$ smallest groups, then $\forall t\in [0,1]$ there exists an $(r,t)$-accordant clustering iff $k\in \{1,...,N- \sum_{i=1}^r \lceil tn_i\rceil+r\}$.
\end{lem}
\begin{proof}
If $k\in \{1,...,N- \sum_{i=1}^r \lceil tn_i\rceil+r\}$, we have two cases either $r\ge k$ or $r<k$. If $r\ge k$, we can form $k$ clusters with $\lceil tn_i\rceil$ instances from the smallest $k$ groups where $i\in \{1,...,k\}$, and place the remaining instances in the $k^{th}$ cluster, thus obtaining a feasible clustering. If $r<k$, we can form $r$
clusters again with $\lceil tn_i\rceil$ instances from the $r$ smallest groups. With the remaining $N-\sum_{i=1}^r \lceil tn_i\rceil$ instances we can perform $k-r$ unsupervised clustering, since $k-r \le (N- \sum_{i=1}^r \lceil tn_i\rceil+r)-r = N-\sum_{i=1}^r \lceil tn_i\rceil$. This again leads to a feasible accordant clustering.

If $k>N- \sum_{i=1}^r \lceil tn_i\rceil+r$, then to have a feasible clustering firstly, the maximum number of clusters we can have with the instances not required to be in accordance is $N-\sum_{i=1}^r \lceil tn_i\rceil$. So with the remaining instances $\sum_{i=1}^r \lceil tn_i\rceil$ we need to form $k-(N- \sum_{i=1}^r \lceil tn_i\rceil)>r$ clusters. Having to distribute $\sum_{i=1}^r \lceil tn_i\rceil$ instances into more than $r$ clusters leads to infeasibility as at least one of these $r$ smallest groups will end up being underrepresented in all of the $k$ clusters.
\end{proof}

In general, feasibility is unlikely to be an issue as $k \ll N$ in most real applications.


\begin{figure}[t]
\center
\includegraphics[height=0.28\textwidth]{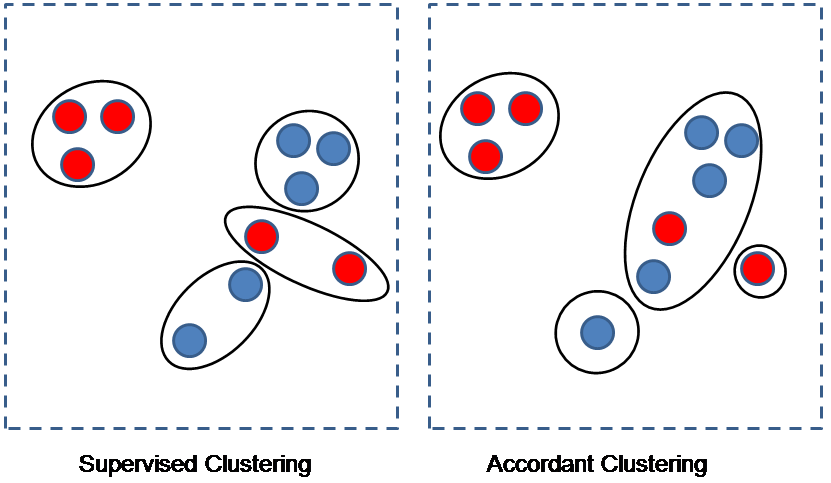}
\caption{Contrast between supervised and accordant clustering, where the groups/labels are showed in two colors. The figure on the left hand side depicts a clustering where each cluster has homogeneous labels, which is one of the main objectives of supervised clustering. However, this clustering is \emph{not} accordant (for $t=0.75$), since no cluster contains at least 75\% of any group. On the other hand, the figure on the right shows an accordant clustering that is not as desirable from a supervised clustering perspective, due to the presence of a cluster with mixed labels. This illustrates the supervised and accordant clustering have distinct objectives, where neither paradigm is strictly stronger than the other.}
\label{ACeg2}
\end{figure}

\subsection{Contrast with Other Frameworks}~\label{comparison}
In the traditional clustering paradigm, the goal is to partition data into (meaningful) clusters. To this end, a wide variety of objective functions and algorithms have been proposed, falling into a fairly large number of distinct clustering paradigms, which  vary in the types of input and output required for clustering methods \cite{charubook}. 

The most fundamental paradigms are either partitional, where the output is a set of $k$ (disjoint) clusters or hierarchical, which simultaneously represents multiple partitionings in a tree structure. Another popular variation is soft clustering, where points may belong to multiple clusters, compared with the classical hard clustering model where each point is part of a unique cluster. Naturally, variations in how the output is represented offers no way of representing from which groups elements derive, and as such does not aid in the discovery of group insights. 

The type of input that different clustering techniques accept is remarkably wide. One such variation allows the user to specify a weight~\cite{weight} representing the significance of individual elements, which guides the clustering in terms of which instances should be given more importance. While allowing more flexibility, a weight is associated with a particular instance, which does not enforce any condition on groups of instances to be assigned to the same cluster.

Perhaps the most popular semi-supervised setting allows certain pairs of instances to be marked as must-link (ML) or cannot-link (CL) \cite{wagstaff,ian}. If such constraints are feasible~\cite{ianfeasibility}, the final clustering is likely to have semantic value that is of use to the practitioner. If the data is partially labeled, the goal often becomes to attain an optimal objective cost while respecting the labeling. In the extreme case of supervised clustering \cite{super,thorsten,skmeans} the entire dataset is labeled. Note that the partially/fully labeled settings could be modeled as pairwise constraints and the machinery used for constraint based clustering could be used in this case too, though the number of constraints could potentially be quadratic in number of labeled examples. In both cases though, the goal is to produce a clustering that is more or less consistent with the labeling or pairwise constraints.




Observe that unlike supervised clustering, accordant clustering does not imply that the clusters should be homogeneous with respect to the labels, where the groups could be considered as proxies for class labels, but rather \emph{a large fraction of instances belonging to some group should be present in some cluster}. This does not penalize a cluster containing a sizable number of instances belonging to other groups. In fact, if 2 or more groups have $>t$ fraction in the same cluster this could lead to a unified action across the groups, which could be highly favorable.

Moreover, a clustering which is excellent from the supervised perspective may not be feasible relative to our constraint, as each cluster may be homogeneous and contain only a single group and yet no cluster may contain at least $t$ fraction of the instances from any specific group. An example of this is seen in figure~\ref{ACeg2}. Given 4 clusters with $t=0.75$ as before, the clustering on the left has no impurity and is an excellent clustering from the supervised clustering perspective. However, it is not accordant, since a consistent strategy is hard to put into place at the group level, given that the instances are spread across different clusters. 

In particular, the spread implies a lack of cohesion within and amongst any of the groups, making it difficult for practitioners to qualitatively interpret the groups based on the clustering. The clustering on the right is what would be reasonable in our setting. In the cases that supervised clustering does satisfy our constraint, we might see that it is an overkill as it unnecessarily devalues the unsupervised clustering objective giving us a much worse clustering since, it strives to enforce homogeneity across all clusters. We will see evidence of this in the experimental section.

Our definition of usefulness cannot be effectively captured in the constraint-based or label-based semi-supervised clustering frameworks either. The reason being that we do not know which $t$ fraction of the instances belonging to a group should be assigned to some cluster, so as to obtain a high quality clustering. 

It could be argued that we could randomly choose these instances and then perform semi-supervised clustering. However, we might have missed a different set of instances which if we had chosen as the $t$ fraction, would have resulted in a much better clustering. Thus, if we knew this better set a priori, then we could model it with ML constraints or assign its instances the same label. Unfortunately, we do not and hence, the clustering algorithm needs to find this set - in fact, finding this set is one of the main objectives of an accordant clustering algorithm. 

Our algorithm in section \ref{meth}, performs this task and can be shown to converge. Our framework therefore requires the dynamic identification of instances from a group that will lie in the same cluster, which is not the case for the semi-supervised framework.


\textbf{Why cluster all the data?} Our goal, as mentioned before, is to understand relationships within and between groups, so the latter would be lost if we clustered each group independently. For instance, consider the application where we  try to discover over/under performing schools. The good students of one school may turn out to be under-performing students when considering all schools. So independently clustering would not readily provide the necessary insight. Moreover, independently clustering does not make the problem computationally easier as cardinality constraints are hard to solve. \textit{Note however that one can always use only the relevant data to perform accordant clustering by removing groups that are known to be unimportant operationally or possibly because they are too small}.

Our framework is also different from subgroup discovery \cite{subgrp}, which mainly tries to find rules in conjunctive form relative to a given target. Our goal is not to find rules that lead to certain characteristics of the target, but rather to find insights about the groups based on the inputs by having the groups well represented in clusters. These insights could be across groups and not distinct for each group. Moreover, there is no restriction in having only conjunctions as at least $r$ groups from $m$ may be well represented in the same or different clusters which can also lead to disjunctions.

\section{Accordant $k$-means}~\label{meth}
Given that $k$-means may be the most frequently used clustering paradigm, it serves as a natural foundation for developing an accordant clustering technique. Algorithm \ref{algo1} aims to minimize the sum of squares error (SSE), i.e. the $k$-means objective. The distinguishing feature of this algorithm lies in its effort to satisfy the accordant constraint by considering the penalties associated with sub-optimal point to center assignments, where the ``penalty'' of assigning point $x_i$ to center $c_j$ is $d(x_i, c_j) - \min_{\ell} d(x_i,c_{\ell})$.


While the accordant $k$-means algorithm (abbreviated Akmeans) is primarily concerned with satisfying its constraint, it does so in a way that minimizes the associated penalty. As such, Akmeans attempts to uncover the accordant clustering which is also optimal w.r.t. its objective function.


\begin{algorithm}[h]
    \caption{Accordant $k$-means (Akmeans)}
    \label{algo1}
\begin{algorithmic}
\State Choose $k$ random centers $\{c_1,~\ldots,~c_k\}$ from $X$
\State {\bf repeat until} convergence:
    \State\tab  For each $x_i \in X$:
    \State\tab\tab  For each cluster center $c_j$, compute penalty \tab\tab $\mathcal{P}_{ij}$$= d(x_i, c_j) - \min_{\ell} d(x_i,c_{\ell})$
	\State \tab For each group $X_j$ and each center $c_i$:
	\State\tab \tab Sort the points of $X_j$ in ascending penalty.
	\State\tab \tab  The sum of penalties for the first $t$ fraction \tab\tab of  points is the penalty of this pairing.
    \State\tab Choose the $r$ lowest penalty group-center pairs.
	\State\tab  Assign first $t$ fraction of points in these chosen \tab pairings to the corresponding cluster centers.	
	\State\tab  Assign remaining points to the closest cluster \tab center.	
	\State\tab  Compute new cluster centers $\{c_1,~\ldots,~c_k\}$.
    \State Output final clustering $\mc{C}$ based on latest centers.
\end{algorithmic}
\end{algorithm}



\begin{figure}[t]
\center
\includegraphics[height=0.2\textwidth]{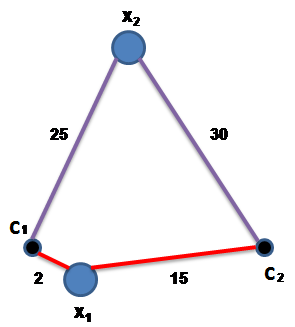}
\caption{Above we see the Euclidean distance squared of instances $x_1$ and $x_2$ from cluster centers $c_1$ and $c_2$.}
\label{ACp}
\end{figure}

\subsection{Description}

Our method takes as input the groups and the fraction $t$, besides the standard inputs to $k$-means. $\tau$ and $\delta$
can be used to specify termination conditions, where $\tau$ is the maximum number of allowed iterations, while $\delta$ is the maximum difference between the objective function at successive iterations at or below which we claim convergence. We may choose both or either of these conditions to indicate termination of algorithm
\ref{algo1}.

The crux of the algorithm and where it differs from standard $k$-means is that it has to choose $t$ fraction of the instances belonging to $r$ groups that must lie in up to $r$ clusters at each iteration. This implies that every intermediate clustering based on our algorithm in the path to convergence is also feasible.

As in standard $k$-means, in this case too we compute a $N\times k$ distance matrix $\mathcal{D}$, which stores the \emph{squared} Euclidean distances between instances and the current cluster centers. However, for Akmeans we compute another $N\times k$ matrix called the penalty matrix $\mathcal{P}$, which computes the penalty of assigning an instance to a specific cluster. The penalty of assigning an instance $x_i$ to cluster $C_j$ is given by,
$\mathcal{P}_{ij}=\mathcal{D}_{ij}-\min_{s\in \{1,...,k\}} \mathcal{D}_{is}$.

Consequently, if $c_j$ the cluster center of $C_j$ is the closest cluster center to $x_i$, then $\mathcal{P}_{ij}=0$. It is greater than zero for farther away clusters. Thus, $\mathcal{P}_{ij}$ $\forall j\in\{1,...,k\}$ is the excess amount that would be added to the clustering objective if $x_i$ is assigned to $C_j$
rather than to its closest cluster during the current assignment step. In our algorithm, we try to choose $t$ fraction of the instances belonging to a particular group along with their assignment to a specific cluster such that the sum of their penalties is minimum. Hence, for each group and cluster we select $t$ fraction of the instances belonging to that group with the lowest penalties and compute their sum. For $r$ group-center pairings with the lowest penalty we assign these instances to the corresponding clusters. The remaining instances are assigned as in standard $k$-means to the closest cluster. We then compute the means of these new clusters and iterate through the above two steps until one of the termination conditions is reached.

At each iteration, it is better to choose the $t$ fraction of the instances based on the penalty matrix than the distance matrix, since we are deviating the least from the unconstrained version relative to the attained objective value. If we were to choose the instances based on the minimum sum of the distances of $t$ fraction of the instances belonging to a group, then we may not achieve reduction in objective value to the extent possible during that iteration. A simple illustration of this is seen in figure \ref{ACp}. If we are to assign one of the two instances $x_1$ or $x_2$ to $c_2$, then based on squared distances we would assign $x_1$ to $c_2$ since $x_1$ is closer to $c_2$ than $x_2$ is to $c_2$. With this assignment the objective value has increased by $15-2=13$ over the objective based on unsupervised clustering. However, if we assign based on our strategy of minimum penalty, then $x_2$ would be assigned to $c_2$ rather than $x_1$. This is so, since the penalty for $x_1$ is $13$, while the penalty of assigning $x_2$ to $c_2$ is just $30-25=5$. Thus, the objective value would now be worse of by $5$, rather than $13$
relative to the unsupervised clustering objective. This all is of course because unsupervised clustering would assign both the instances to $c_1$.

\subsection{Convergence and Time Complexity}

\eat{
While Akmeans is primarily concerned with satisfying its constraint, it does so in a way that minimizes the associated penalty. As such, Akmeans attempts to uncover the accordant clustering which is also optimal w.r.t. its objective function. 
}
We now show that, like the traditional $k$-means algorithm, our algorithm provably converges.

\begin{lem}\label{convergence}
The Akmeans algorithm converges.
\end{lem}
\begin{proof}
Since there are only a finite number of partitions of a dataset of size $N$, to prove convergence, it suffices to show that the objective is monotonically decreasing with each iteration.

At any iteration $i$ our algorithm produces a feasible clustering $C_i$. Now at iteration $i+1$ we could maintain the $r$ sets of $t$ fraction assignments as they are and only assign the remaining points to closest centers. Let us denote this clustering by $C_{i+1}^r$. This will reduce or maintain the cost i.e. if $\Phi$ is the objective function $\Phi(C_i)\ge \Phi(C_{i+1}^r)$. However, our method at iteration $i+1$ considers $C_{i+1}^r$ as one possible alternative and chooses an assignment that is no more than $\Phi(C_{i+1}^r)$. Hence, $\Phi(C_i)\ge \Phi(C_{i+1})$. The last step of recomputing the centers further reduces or maintains the cost thus proving that our algorithm produces a monotonically decreasing sequence.
\end{proof}

The time complexity per iteration of $k$-means is $O(Nk\rho)$, where $N$ is the dataset size and $\rho$ is the dimensionality. If $n_{max}$ is the size of the largest group, then the complexity of our method is $O(mn_{max}log(n_{max})k\rho)$. The extra $log$ factor comes from having to sort the penalties of datapoints in each group $k$ times.

\section{Qualitative Guarantees}

As is the case for any clustering paradigm, an arbitrary partitioning of a dataset is not meaningful \emph{a priori} and as such, it is critical to identify clusterings which reveal some inherent structure in a dataset. In addition to revealing this structure, an accordant clustering must have some clusters that are comprised of a substantial proportion of at least one of the groups. Hence, the goal in this framework is to find the accordant clustering of a dataset that best represents its natural structure. In this section, we prove that Akmeans is opt at discovering high quality accordant clusterings when they are present in the data. 
\eat{
A wide range of notions of \emph{clusterability} have been developed which qualitatively express the extent to which a dataset exhibits clustering structure \cite{ackerman2009clusterability}. Several of these notions can be used to characterize the types of structure that optimize certain objective functions. However, despite an attempt to describe the same tendency, many notions of clusterability are pairwise inconsistent \cite{ackerman2009clusterability}. As such, these notions must be invoked with some care w.r.t. the application at hand.}

One of the most insightful and widely-used notions of clusterability related to the $k$-means objective function is the $(c,\epsilon)$-property \cite{ackerman2009clusterability, bbg:approx, Ostrovsky06}, which describes a dataset characterized by a unique clustering that optimizes the objective (see Balcan \emph{et. al}~\cite{bbg:approx} for a detailed exposition). 

\begin{figure}[tb]

\centering
	\includegraphics[scale=0.3]{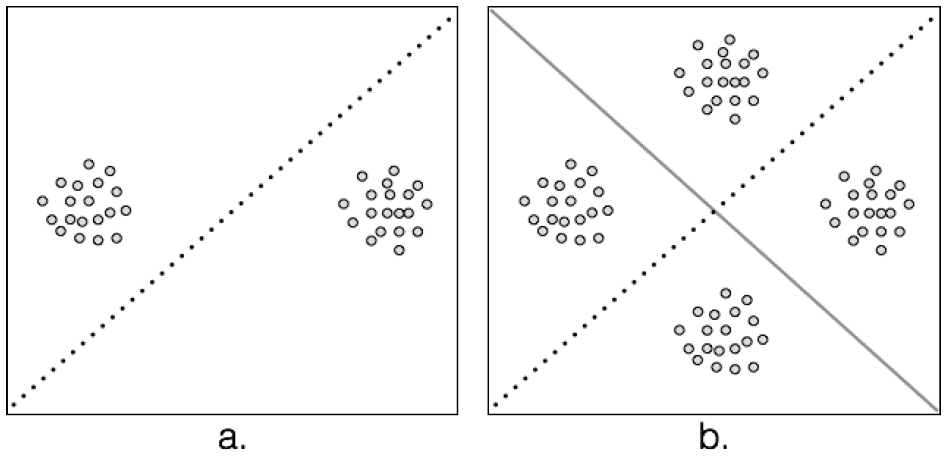}
\caption{The dataset depicted in (a) satisfies the $(c,\epsilon)$-property for the $k$-means objective when $k=2$. The optimal partitioning is shown by the dividing line. Clearly, any near optimal clustering will by necessity approximate this partitioning. Conversely, (b) depicts data which fails to satisfy the $(c,\epsilon)$-property. As shown, there are two radically different ways to partition the dataset for $k=2$ in a way that optimizes the objective.}
\label{fig:uo}

\end{figure}

Intuitively, this property reflects a dataset which has an optimal clustering that is unique in the sense that any clustering of similar cost must be structurally similar to the optimal, as depicted in Figure \ref{fig:uo}. Given two $k$-clusterings $\mc{C} = \{C_1, \ldots, C_k\}$ and $\mc{C'} =\{C'_1, \ldots, C'_k\}$, let $dist(\mc{C},\mc{C'})$ be the fraction of points on which they disagree under the optimal matching of clusters in $\mc{C}$ to clusters in $\mc{C'}$, that is, $dist(\mc{C},\mc{C'})= min_{\sigma \in S_k} \frac{1}{n}\sum_{1}^k|C_i-C'_{\sigma(i)}|$, where $S_k$ is the set of bijections from $[k]$ to $[k]$.

\begin{definition}[{$(c,\epsilon)$-property}~\cite{bbg:approx}] \label{ceuo}
A dataset $(X,d)$ satisfies the $(c,\epsilon)$-property for objective function $\Phi$ if for every $k$-clustering $\mc{C}$ of $X$ where $\Phi(\mc{C}) \leq c\cdot \tup{OPT}_{\Phi}$, the relation $dist(\mc{C},\mc{C}^{*}) < \epsilon$ holds, where $\mc{C}^{*}$ is the clustering that optimizes the value of $\Phi$.
\end{definition}

We show that when a data is clusterable w.r.t. the above notion, and contains an accordant clustering of near-optimal cost, then it can be uncovered within a small number of points.

The cores of a clustering represent a small set of points in each cluster for which every other point in the cluster is closer to than to data outside the partition. 


\begin{definition}[Core]
For any clustering \\ $\mathcal{C} = \{C_1, \ldots, C_k\}$ of  $(\mathcal{X},d)$, the \emph{\bf core} of cluster $C_i$ is the maximal subset $C_i^o \subset C_i$ such that $d(x,z) < d(x,y)$ for all $x \in C_i$, $z \in C_i^o$, and $y \not\in C_i$.
\end{definition}
The proofs for Theorem 1 and Corrolary 1 are in the supplementary material\footnote{The supplementary material is on the first authors website.}.

\begin{thm}\label{structured_data}
Let $(X,d)$ be a dataset which satisfies the $(\alpha, \epsilon)$-property that contains a near-optimal $(r,t)$-accordant clustering with cluster cores of size at least $\epsilon' n$. Then Akmeans outputs an $(r,t)$-accordant clustering that is $2\epsilon$-close to the optimal $(r,t)$-accordant clustering $C_A$ with probability at least $1 - ke^{-\epsilon' k}$.
\end{thm}

The following corollary extends Theorem \ref{structured_data} across multiple initializations.

\begin{cor}\label{structured_data:cor}
Let $(X,d)$ be a dataset which satisfies the $(\alpha, \epsilon)$-property that contains a near-optimal $(r,t)$-accordant clustering with cluster cores of size at least $\epsilon' n$. If Akmeans is run $m$ times, selecting the lowest cost clustering, we find an $(r,t)$-accordant clustering that is $2\epsilon$-close to the optimal $(r,t)$-accordant clustering with probability $1 - (ke^{-\epsilon' k})^m$.
\end{cor}


\eat{
\begin{table*}[htbp]
\begin{center}
  \begin{tabular}{|c|c|c|c|c|c|c|c|}
    \hline
$k$ & $k$-means & Akmeans & Skmeans & SSIkmeans & COPkmeans & CSC & GFHF \\
\hline
\hline
$2$ & 513.5 & 513.5 & 482.3 & 6492.5 & 22199.8 & 16595.7 & 37903.4\\
\hline
$3$ & 297.2 & 1999.2 & 1571.5 & 5973.5 & 19225.3 & 16112.2 & 45340.9\\
\hline
$4$ & 117.8 & 2244.6 & 3867.8 & 6173.8 & 23943.1 & 17581.5 & 48660.2\\
\hline
$5$ & 100.2 & 2148.9 & 4117.1 & 5292.6 & 25362.8 & 26589 & 50149.1\\
\hline
$6$ & 520.5 & 1574.9 & 2470.3 & 5382.3 & 25700 & 16742 & 48677.4\\
\hline
$7$ & 420.7 & 2177.1 & 3756.7 & 4836.2 & 25234.7 & 11883.6 & 50216.9\\
\hline
\end{tabular}
\end{center}
  \caption{The above table shows half the width of the 95\% confidence interval (based on the randomizations) for the different methods and for different values of $k$ around the corresponding means w.r.t. the Spend dataset.}
\label{spendtab}
\end{table*}
}


\begin{figure}[t]
\begin{center}
\begin{tabular}{ cc }
	\includegraphics[scale=0.2]{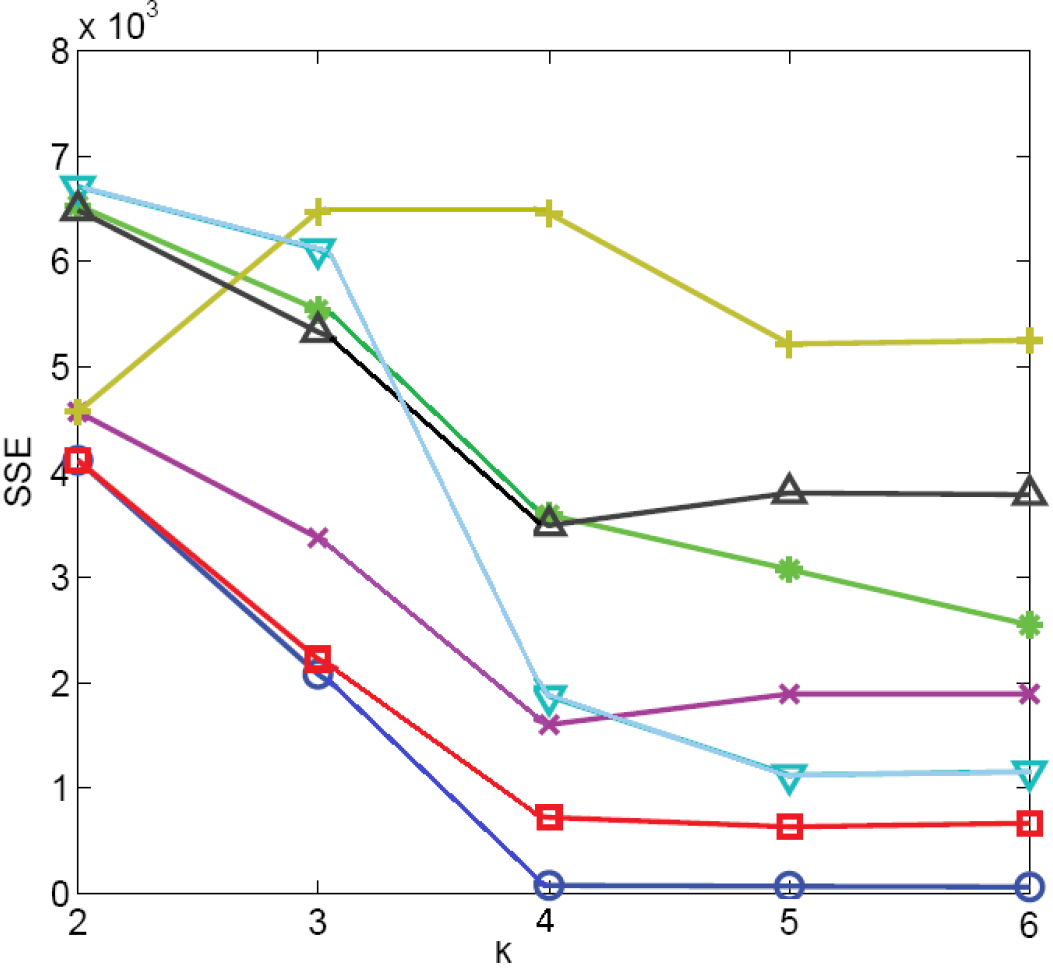}
	& \includegraphics[scale=0.2]{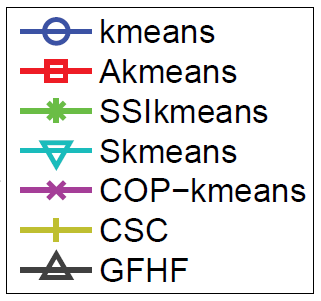}\\
\end{tabular}	
\end{center}
\caption{Above we see the (mean) performance of the different methods with varying $k$ for $t=0.75$ on the proprietary Health Care dataset. $k$-means does \emph{not} satisfy our constraint. The 95\% confidence intervals are given in Table 1 in the supplementary material.}
\label{realmed}
\end{figure}

\eat{
\begin{table*}[htbp]
\begin{center}
  \begin{tabular}{|c|c|c|c|c|c|c|c|}
    \hline
$k$ & $k$-means & Akmeans & Skmeans & SSIkmeans & COPkmeans & CSC & GFHF \\
\hline
\hline
$2$ & \small{41.2} & \small{41.2} & \small{88.3} & \small{749.9} & \small{2519.4} & \small{1899.2} & \small{3212.4}\\
\hline
$3$ & \small{31.5} & \small{21.4} & \small{183.5} & \small{698.8} & \small{2245.3} & \small{1834.2} & \small{39344.9}\\
\hline
$4$ & \small{9.5} & \small{22.5} & \small{456.3} & \small{677.7} & \small{2698.1} & \small{1829.5} & \small{4171.9}\\
\hline
$5$ & \small{10.2} & \small{21.3} & \small{491.1} & \small{594.6} & \small{2746.8} & \small{1704.4} & \small{42840.1}\\
\hline
$6$ & \small{13.5} & \small{57.6} & \small{258.5} & \small{619.6} & \small{2734.4} & \small{1840.4} & \small{4690.4}\\
\hline
\end{tabular}
\end{center}
  \caption{Above we see half the width of the 95\% confidence interval (based on the randomizations) for the different methods, for different values of $k$ around the corresponding means w.r.t. the Health Care dataset.}
\label{hlth}
\end{table*}
}
\section{Experiments}\label{exp}

When applied in practice, on two separate occasions the notion of accordant clustering resulted in insights which domain experts acted upon and those that couldn't be readily found by other methods. The first instance is in the field of medicine, using a dataset representing patients who suffered from a neurodegenerative disease, each belonging to one of five distinct treatment groups depending on the care they received. The second instance was in the field of business, for a Spend dataset representing two years of transactional data from a large corporation, with each transaction falling into one of 25 categories (such as IT, Research, Marketing, etc.). Moreover, we also perform experiments on 6 UCI datasets showcasing the power of our method in uncovering higher quality accordant clusterings.

These datasets provide a point of comparison for measuring the quality of clusterings obtained by the Akmeans algorithm relative to several other state-of-the-art methods, both supervised and semi-supervised, which were adapted to this setting and prepared for these datasets so as to have a fair comparison. Specifically, the methods chosen for comparison are as follows: 1) Supervised $k$-means (Skmeans) \cite{skmeans}, 2) SVM based supervised iterative $k$-means (SSIkmeans) \cite{ssikmeans}, 3) Constrained $k$-means (COPkmeans) \cite{wagstaff}, 4) Constrained spectral clustering (CSC) \cite{kkm,xiang} and
5) Semi-supervised learning based on Gaussian fields and harmonic functions (GFHF) \cite{GFHF}.

The quality of the clustering is measured by the SSE, as a majority of these methods are extensions of the $k$-means algorithm with the others known to be competitive relative to this metric. Additionally, the performance of standard $k$-means is included to act as a baseline for the SSE achieved by all other methods.
\emph{Similar qualitative results were observed using other measures, such as mutual information, silhouette and Davies-Bouldin index}. For each of these methods we set $\delta=10^{-7}$ to detect convergence. 

\emph{Note that, since we cannot implicitly enforce satisfaction of our accordant constraint for each initialization of all algorithms except for  Akmeans, the reported results represent an average over runs which resulted in accordant clusterings}. Thus, the SSE considered for each of the methods accounts only for clusterings that satisfied the accordant constraint, and consequently we would prefer algorithms that provide tight clusters with low SSE.

\begin{figure}[t]
\begin{center}
\begin{tabular}{ cc }
	\includegraphics[scale=0.2]{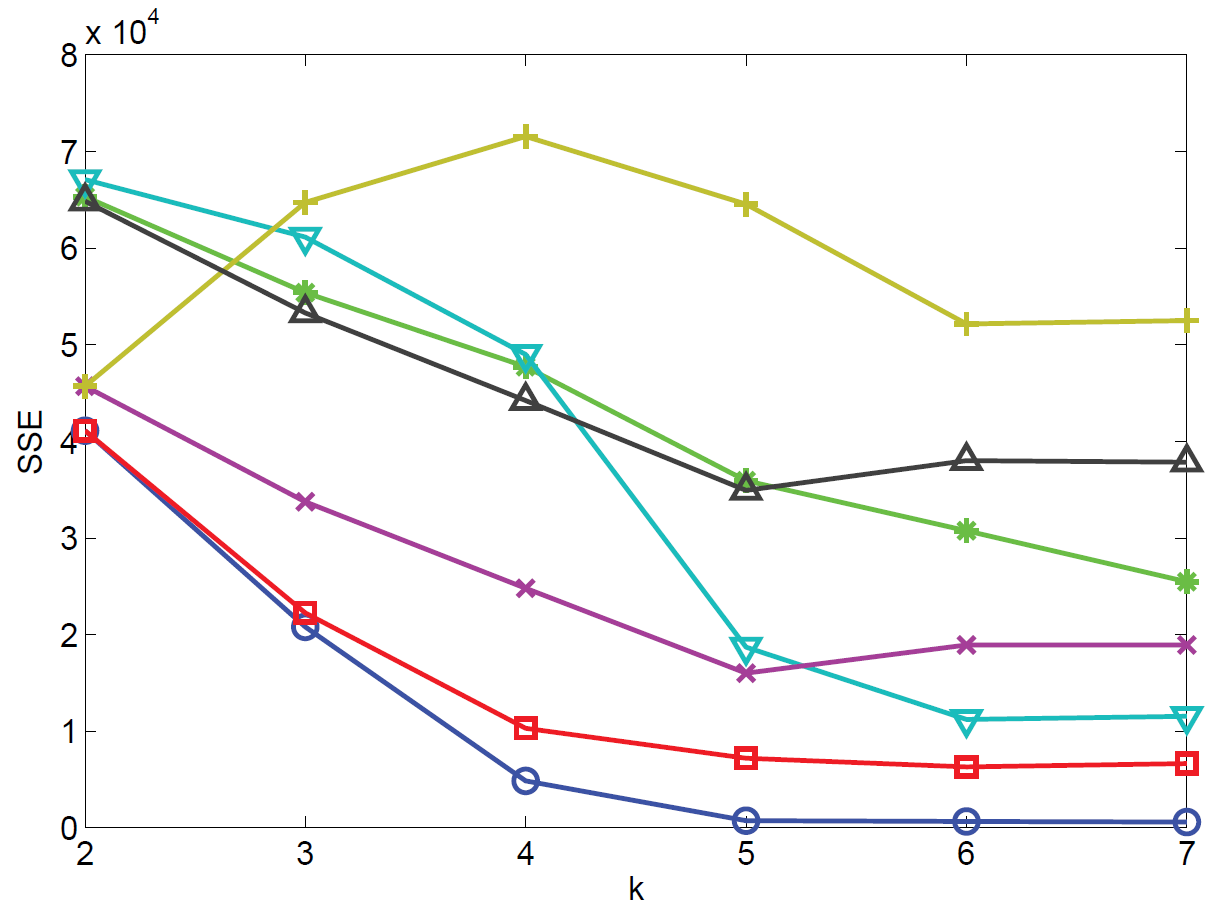}
	& \includegraphics[scale=0.2]{graph_labels}\\
\end{tabular}
\end{center}
\caption{Above we see the (mean) performance of the different methods with varying $k$ for $t=0.8$ on the proprietary Spend dataset. $k$-means does \emph{not} satisfy our constraint. The 95\% confidence intervals  are given in Table 2 in the supplementary material.}
\label{realspend}
\end{figure}

We have two supervised methods Skmeans and SSIkmeans. For Skmeans we set a high weight for the group number attribute so that it satisfied our constraint even for high values of $t$. SSIkmeans was implemented by installing the python interfaces \cite{thorsten} to SVM-light \cite{svmlight}. We use the iterative variant rather than the spectral one, such that it is under-constrained for supervised clustering and hence, should yield better quality clusterings when the accordant constraint is satisfied. The models were trained on a random selection of $k$ groups from the dataset, so that each could be applied to the entire dataset to obtain the desired $k$-clustering for each trial. The results were then averaged over $100$ such randomizations. This procedure is necessary for SSIkmeans, since this method implicitly assumes that the number of clusters is equal to the number of different groups that it is trained on.

The remaining three methods COPkmeans, CSC and GFHF are semi-supervised. COPkmeans and CSC are constraint-based semi-supervised clustering methods, which are prepared by randomly selecting some $t$ fraction of the instances belonging to $r$ randomly chosen groups, and assign ``must-link'' (abbreviated ML) constraints to them. COPkmeans incorporates these ML constraints into the $k$-means objective. CSC, on the other hand, modifies the graph affinity matrix based on these ML constraints and then performs spectral clustering on the modified graph.

The GFHF method is a label-based semi-supervised approach. Again, we randomly choose $t$ fraction of the instances belonging to some randomly chosen $r$ groups, but here we assign the same label (which may be the group number) to the corresponding instances rather than adding constraints. We also randomly choose a small fraction ($\approx$5\%) of the instances from other groups, where each small fraction belongs to a different group and hence has a different label. In all of our experiments, the number of clusters $k$, is bounded by the number of groups, $m$, allowing us to appropriately initialize this method. Such an initialization will result in GFHF outputting a $k$ partition. For these 3 approaches, we average the results over multiple ($100$) such randomizations.

For the graph-based approaches (viz. CSC and GFHF), we constructed the graphs using a radial basis kernel after standardizing the data. For $k$-means and its variants, which require initial cluster centers, we randomly choose them such that they all belong to different groups. Thus, when $k=m$ we have exactly one instance randomly chosen from each group to be a cluster center. When $k<m$ we have cluster centers randomly chosen from $k$ different groups. We report the results averaged over multiple ($100$) such initializations. 


\subsection{Health Care dataset}
The health care dataset contains demographic and clinical information from $5,022$ patients who suffer from a neurodegenerative disorder. These patients were divided into 5 (nearly) equisized groups based on the treatment they received. 

Each patient was represented by 57 attributes capturing individual information such as gender, age, race, cognitive decay, and physical condition in addition to treatment specific information such as duration, as well as multiple attributes indicating whether a certain chemical/medicine was used and the corresponding dosage. Given all of these factors, the goal is to identify one or more treatments that may be consistently either effective or ineffective based on the patients cognitive and physical condition. Such information can be a big step towards creating a successful cure for the disease.

Hence, from a modeling perspective, we have $m=5$ groups. After speaking to the medical professionals, it was decided that at least 75\% of a group should be represented in some cluster, i.e. $t=0.75$. The results from the clustering of the different methods for multiple values of $k$ are seen in Figure \ref{realmed}. Our method with $r=1$ and $r=2$ yielded the same results.

We see from the figure that the unsupervised objective, which does \emph{not} produce accordant clusterings for $k>1$, flattens out around $k=4$. This suggests that there are probably 4 clusters in this dataset. We observe that Akmeans, which produces accordant solutions, is the closest in its SSE to traditional $k$-means. It is in fact significantly better than its (adapted) competitors.

In the Akmeans clustering at $k=4$, we observed that treatment groups $1$ and $3$ ended up satisfying our constraint and in fact lying within the same cluster. This cluster was characterized by much better patient condition relative to other clusters. Additionally, \emph{this result proved to be particularly interesting since, the mean/median conditions computed across each (entire) group are practically indistinguishable}. The medical professionals were quite excited by this finding and have decided to:
\begin{enumerate} \itemsep2pt \parskip0pt \parsep0pt
\item Perform further studies specifically focused on the treatments given to groups $1$ and $3$.
\item Begin administering treatments corresponding to groups $1$ and $3$ to a wider pool of patients in time.
\end{enumerate}

\subsection{Spend dataset}

The Spend dataset contains a couple of years worth of transactions spread across various categories belonging to a large corporation. There are $145,963$ transactions which are indicative of the companies expenditure in this time frame. The dataset has 13 attributes, namely: requester name, cost center name, description code, category, vendor name, business unit name, region, purchase order type, addressable, spend type, compliant, invoice spend amount. Given this the goal is to identify spending and/or non-compliant tendencies amongst one or more of the 25 categories. With this information, the company would then be able to put in place appropriate policies and practices for the identified categories that could lead to potentially substantial savings in the future.

Hence, we have $m=25$ groups in our dataset. With the help of domain experts, it was decided that at least 80\% of the transactions belonging to a single category should exhibit a similar tendency or pattern in order for them to consider taking any action. Consequently, we set $t=0.8$. The results from the clustering of the different methods for multiple values of $k$ are seen in Figure \ref{realspend}.

We see from the figure that the unsupervised objective, which does \emph{not} produce accordant clusterings above $k=2$, flattens out more or less at $k=5$. This suggests that there are probably 5 true clusters in the dataset. We observe that Akmeans, which produces accordant clusterings, is again the closest in performance to traditional $k$-means at $m\geq k$.

In the Akmeans clustering at $k=5$, we observed that the constraint was satisfied for the marketing category. The corresponding transactions for this category were characterized by high spend that was mostly non-compliant with company guidelines. This type of insight can be very useful for a company as, once aware of this spending, it can focus its efforts on this particular category rather than spread itself thin by expending effort across multiple areas. In fact, based on a review of these results with domain experts they acknowledged that this was indeed insightful and could lead to the following actions:
\begin{enumerate} \itemsep2pt \parskip0pt \parsep0pt
\item Stricter monitoring of travel expenditure of employees in marketing.
\item Tighter controls and extra approvals for marketing campaigns and advertisements that have expenditures exceeding a certain amount.
\item Close monitoring of spend with certain vendors.
\end{enumerate}

\subsection{UCI datasets}

We also evaluated our methods on 6 UCI datasets used in previous clustering studies \cite{xiang} namely: a) Glass, b) Heart, c) Ionosphere, d) Breast Cancer, e) Iris and f) Wine. 

The results are depicted in figure 1 in the supplementary material. We consistently see across all the datasets that Akmeans matches the performance of $k$-means when our constraint is trivially satisfied, while providing statistically significant lower error clusterings than its adapted competitors in other cases.

\section{Discussion}
\label{disc}
In this paper, we introduce a novel clustering paradigm for the discovery of group-level insights. We propose an algorithm based on the $k$-means method that outputs accordant clusters as well as provably uncovers near-optimal solutions on clusterable data. Moreover, we described two real world settings where our algorithm significantly outperformed its adapted competitors, as well as provided actionable insight. Our algorithm's superior performance was further validated by experiments on the 6 UCI datasets. In all cases, our method converged in less than $20$ iterations.

Given the novelty of our framework, here is a realm prime for innovation -- particularly in exploring new applications that can benefit group-level insight. Additional information on cost or penalties may also be incorporated to enable informed action, and our constraint would act as a starting point on top of which additional constraints or regularization terms may be added. A variety of algorithmic challenges may also be addressed in the future, such as exploring methods that discover accordant clusterings with respect to alternate cluster structures and objective functions. One may also try to design a metric that respects our constraint, however, this is far from obvious as we do not want to penalize impurity of clusters but at the same time have $\ge r$ groups well represented in one or more clusters.


\section*{Acknowledgement}
We would like to sincerely thank Joydeep Ghosh, Kiri Wagstaff and Charu Aggarwal for helpful suggestions.

\bibliography{ACRef}  
\bibliographystyle{abbrv}

\end{document}